%% file: ASP.tex
\documentclass[runningheads]{llncs}
\usepackage{graphicx}

\fontfamily{ptm}\selectfont
\usepackage[latin1]{inputenc}
\usepackage[T1]{fontenc}
\usepackage{times}
\usepackage{courier}

\usepackage{amsmath}
\usepackage{amsfonts}
\usepackage{amssymb}

\usepackage{cite}

\usepackage{url}
\usepackage{mathptmx}
\usepackage{wrapfig}
\usepackage{xspace}
\usepackage{paralist}

\usepackage[ruled]{algorithm2e}
\usepackage{breqn}

\usepackage{listings}
\usepackage{float}
\usepackage{color, colortbl}
\definecolor{Gray}{gray}{0.9}
\usepackage{booktabs,caption}
\usepackage[flushleft]{threeparttable}
\usepackage{multirow}
\usepackage{array}
\usepackage{wrapfig}
\usepackage{placeins}

\newcommand{\codett}[1]{\texttt{#1}}

\newcommand{\eat}[1]{}
\newcommand{\prolog}{Prolog\xspace}

\newcommand{\eg}{e.\,g.,\xspace}
\newcommand{\ie}{i.\,e.,\xspace}

\newcommand{\cf}{cf.\xspace}

\usepackage{tikz}
\usepackage{tikzscale}
\usetikzlibrary{shapes,arrows}
\usetikzlibrary{arrows.meta}
\usetikzlibrary{positioning}
\usetikzlibrary{fit}
\usetikzlibrary{shapes.geometric, arrows}
\usetikzlibrary{backgrounds}

\tikzstyle{decision} = [diamond, draw, fill=blue!20, 
    text width=4.5em, text badly centered, node distance=cm, inner sep=0pt]
  \tikzstyle{block2} = [inner sep=2mm, rectangle, draw, rounded corners,fill=blue!10]
  \tikzstyle{block3} = [inner sep=2mm, rectangle, draw, rounded corners]    
    \tikzstyle{block} = [inner sep=2mm, rectangle, draw, fill=yellow!10]
\tikzstyle{line} = [draw, -latex']
\tikzstyle{cloud} = [draw, ellipse,fill=green!20, node distance=3cm,
    minimum height=2em]

\tikzstyle{wa} = [inner sep=2mm, rectangle, draw, fill=yellow!20, 
     rounded corners]

\begin{document}

\title{%
   Towards Generating Explanations for ASP-Based Link
   Analysis using Declarative Program Transformations}

\author{%
   Martin Atzmueller\inst{1}\textsuperscript{*} \and
   Cicek G\"uven\inst{1} \and
   Dietmar Seipel\inst{2}\textsuperscript{*}}
\begingroup
\renewcommand*{\thefootnote}{\fnsymbol{footnote}}
\footnotetext{\textsuperscript{*} Both authors contributed equally to this work.}
\endgroup

\authorrunning{Atzmueller, G\"uven, Seipel}

\titlerunning{Towards Generating Explanations for ASP-Based Link Analysis}

\institute{%
Tilburg University, Cognitive Science and
   Artificial Intelligence,\\
   Warandelaan 2, 5037 Ab, Tilburg, The Netherlands\\
   \email{c.guven,m.atzmuller@uvt.nl}\\[.8ex] \and
University of W\"urzburg, Department of Computer Science, \\
   Am Hubland, D~--~97074 W\"urzburg, Germany\\
   \email{\{dietmar.seipel\}@uni-wuerzburg.de}\\
}

\maketitle

\begin{abstract}

The explication and the generation of \emph{explanations}
are prominent topics in artificial intelligence and data science,
in order to make methods and systems more transparent and understandable
for humans.

This paper investigates the problem of \emph{link analysis}, specifically link prediction and anomalous link discovery in \emph{social networks} using the declarative method of Answer set programming (ASP).
Applying ASP for link prediction provides a powerful declarative approach, \eg for incorporating domain knowledge for explicative prediction.
In this context, we propose a novel method for generating explanations --
as offline justifications -- using declarative program transformations.
The method itself is purely based on syntactic transformations
of declarative programs, \eg in an ASP formalism, using rule instrumentation.

We demonstrate the efficacy of the proposed approach, exemplifying it in an application on link analysis in social networks, also including domain knowledge.

\end{abstract}

\section{Introduction}

Explicative approaches, \ie transparent and explainable methods play an
increasingly important role in the artificial intelligence and data science
communities.
General approaches for generating \emph{explanations} in conjunction with
a given method, or with its results are therefore important and relevant
with a broad range of applications.
In this paper, we focus on this problem in the context of logic programming
approaches, in particular for
\emph{answer set programming (ASP)}, \cf \cite{lifschitz2008answer}. 

Specifically, we present a method for generating explanations for results of an
answer set solver using \emph{declarative program transformations}.
For an answer set and its elements,
we construct a trace of its derivation in terms of the applied rules
and ground atoms, providing a justification~\cite{pontelli2006justifications}.
It is important to note that
the presented method, which relies on purely syntactic declarative program
transformations, is in principle not restricted to ASP-based approaches,
but could also be extended to further logic-based methods and theorem provers.

We demonstrate the application of the proposed method on the problem of
ASP-based link analysis, \ie link prediction and anomaly analysis.
Link prediction and (anomalous) link discovery are prominent methods
in \emph{social network analysis (SNA)},
for which we have recently demonstrated the benefits of
applying logic-based, particularly ASP-based approaches~\cite{CA:2019}.
Essentially, link prediction aims to estimate the future link structure in a
(social) network, while anomalous link discovery focuses on the identification
of links in a network that deviate from a given model of normality, \ie from
expectations, or a specifically formalized model.
In a case study, we show the efficacy of the proposed explanation method and
its impact in the SNA domain, focussing on explanations for predicted/anomalous
links. We apply ASP since it allows to specify
interesting structures and patterns in a compact way. Due to its strength
in including background knowledge by facts (and rules), link prediction
approaches can be easily implemented and complemented if such background knowledge
is available, \cf~\cite{CA:2019}.

Our contributions are formulated as follows:
\begin{compactenum}
   \item We propose a novel method for generating explanations on ASP-based
formalisms using declarative program transformations.
   \item We show the implementation of the method within the
Declare~\cite{Declare} software system, targeting the Clingo
system~\cite{DBLP:journals/corr/GebserKKS14} as the applied answer set programming toolkit.
   \item We demonstrate the efficacy of the presented method, \ie its
applicability and benefits of the proposed method in a case study using
ASP-based link analysis for link prediction and anomalous link discovery, and
obtaining respective explanations.
\end{compactenum}

The rest of the paper is structured as follows:
Section~\ref{sec:related} discusses related work. After that,
Section~\ref{sec:method} outlines the proposed method for generating
explanations. Next, Section~\ref{sec:case:study} presents a case study on
ASP-based link analysis. Finally, Section~\ref{sec:conclusions} concludes with
a summary and outlines interesting directions for future work.

\section{Related Work}
\label{sec:related}

Below, we discuss related work on answer set programming, before
focusing on explication and explanation. Finally, we briefly discuss
related approaches on link prediction.

\subsection{Answer Set Programming}

Answer set programming (ASP)~\cite{niemela1999logic} is a declarative problem
solving approach. Given a problem, ASP aims to find one or several possible
solutions; these are the so-called answer sets, \ie all possible sets of facts
that are consistent with the facts stated earlier to the original problem,
\eg~\cite{kaufmann2016grounding, gebser2016modeling}.
ASP is  designed for NP-hard problems and  finds its
applications in large instances of industrial problems,  since it offers a rich
representation language and high performance solvers; some recent applications
are listed in  \cite{falkner2018industrial}. Some examples of ASP solvers that
are considered to be efficient are Smodels~\cite{syrjanen2001smodels}, dlv~\cite{dell2001system},
WASP~\cite{dodaro2013engineering}, Clasp~\cite{gebser2012conflict} and
Clingo \cite{DBLP:journals/corr/GebserKKS14}. Clingo\footnote{Available at:
\url{https://potassco.org/}} itself combines a powerful grounder (Gringo) with
Clasp (for solving) into an integrated system. For ease of use, and due to its
efficiency (\eg~\cite{guyet2018efficiency,schapers2018asp}), we utilized Clingo
in the context of this paper.
We assume that the reader has some background knowledge about ASP.

\subsection{Explication and Computing Explanations}

Recently, the concept of explicative models and approaches has gained a strong
momentum in artificial intelligence and data science,
\eg~\cite{AR:10a,biran2017explanation} -- aiming at transparent, interpretable,
and explainable models in order to make the models and approaches more
understandable to humans, in the idea of computational
sensemaking~\cite{Atzmueller:18:Declare}. First approaches for generating
explanations in the context of link prediction have been discussed
by~\cite{SA:18:BNAIC}. In this paper, we extend on those approaches providing a
specific implementation using ASP.

Furthermore, reconstructive explanations~\cite{wick92a}, also on several
explanation dimensions~\cite{AR:10a}, is an approach that constructs
explanations by tracing back the steps of a system when constructing its
output. In this sense, this forms an important basis of the approach proposed
in this paper, since we construct explanations considering the specific answer
set and rules that have fired, however targeting ASP in particular using a
flexible declarative approach for program transformation. In the ASP-domain
itself, constructing explanations and justifications is also a prominent
topic~\cite{fandinno2019answering}, having emerged in recent
years~\cite{pontelli2006justifications,erdem2016applications,fandinno2019answering}. Debugging techniques for ASP based on rewriting rules have been proposed, \eg in~\cite{brain2007debugging,gebser2007spock}.
Also, justifications and justification
trees~\cite{pontelli2006justifications,fandinno2019answering} of a derived
answer set solution are related to our approach. We apply a similar technique
for generating explanations, however, our method is different (and more
general) in at least two ways: First, we do not only generate justifications,
but extend on those by providing a user-specific selection on the given
knowledge elements. Furthermore, our approach is more general, since we apply
declarative program transformations that only perform syntactic
transformations, and are not necessarily specific for ASP programs.

\subsection{Link Analysis}

Link analysis and mining encompasses several techniques and
methods~\cite{getoor2005link}. In the context of this paper, we focus on link
prediction and the anomalous link discovery.

The focus of link prediction is the dynamics and mechanisms in the creation of
links between the parties in social networks ~\cite{LK:03}. Such networks are
typically represented as graphs, where nodes denote the parties, while edges
model the links, \ie the relationships between those parties. Then, the link
prediction problem can be defined as the search to carefully predict edges that
will be added to a given snapshot of a social network during a given interval,
using network proximity measures of the nodes, \ie based on how close the
different nodes are in terms of their common set of neighbors in the
network/graph. In~\cite{CA:2019} we have presented the application of ASP for link
prediction in social networks. In contrast to this approach, this paper does
not focus on the link prediction by itself, but on the explanation or
justification why a given result set (or a specific atom denoting a link) has
been computed. For that reason, we apply ASP and declarative explication.

Anomalous link discovery~\cite{rattigan2005case} aims at discovering anomalous
links, \eg using link prediction that are not highly likely, or that are in contrast to a
given (reference) model~\cite{CA:2019}. Compared to existing approaches, this
paper does not present a new automatic method or approach for that. Instead, we
present a simple model-based technique formalized using ASP, and show how to
generate explanations for the anomalous links.

\section{Method: Generating Explanations by Program Transformation}
\label{sec:method}

In this section, we describe our proposed method for generating explanations
using declarative program transformations. We first give a bird's eye view on the
proposed approach, before we describe its implementation in detail.

\subsection{Overview}

We provide an overview shown in Figure~\ref{fig:workflow}, which depicts
the workflow for the generation process. Please note, that in the process
workflow, we indicate all steps performed by Declare with a ``D'' in red, while
all processing steps involving Clingo are marked with a ``C'' in blue. We  can
roughly divide the process into two phases, \ie program transformation and
evaluation, as well as explanation generation and presentation.

\subsubsection{Phase 1: Program Transformation and Evaluation by Instrumentation}

\begin{enumerate}
   \item We start with a \emph{Clingo Program} which is processed by
\emph{Declare} into a \emph{Declare Program}. This basically involves simple
syntactic transformations.
   \item The next steps for the more complex program transformation
involves the syntactic transformation of the Declare program into an
\emph{Extended Declare Program} -- by enriching the program as described below, using rule instrumentation.
This program is then converted into an \emph{Extended Clingo Program}.
   \item Finally, this enriched program (including statements that allow
the tracing and explanation generation) is evaluated by Clingo and results in
the \emph{Extended Answer Set}.
It is important to note, that with a simple filtering operation, which
removes all the extensions made by the \emph{Extended Clingo Program} we can obtain
the \emph{Answer Set}, which could also be obtained by direct evaluation
of the \emph{Clingo Program} by Clingo.
\end{enumerate}
    
\subsubsection{Phase 2: Incremental Explanation Generation and Presentation}

\begin{enumerate}
   \item We start with the \emph{Extended Answer Set}, which is processed using
Declare based on the answer set obtained from Clingo.
   \item Based on that result, we can generate the \emph{Explanation} using
Declare. For that purpose, the trace information provided by the enriched program is
utilized by Declare for constructing the explanation in reconstructive fashion.
   \item Finally, the explanation is presented to the user as a
\emph{Presentation}. In this step, there is an incremental feedback loop with
the explanation step, in order to include queries of the user, and to tailor
the explanation and its presentation, respectively, to the context and
interests of the user. For some more advanced queries, the explanation can also
be refined, however, this requires a more comprehensive feedback loop back to
the initial \emph{Declare Program}. Then, this program can be extended and
transformed accordingly to support more sophisticated explanation options like
additional user constraints, domain knowledge, or summarization techniques.
The latter can then be used, for example, for condensing the explanation.
\end{enumerate}

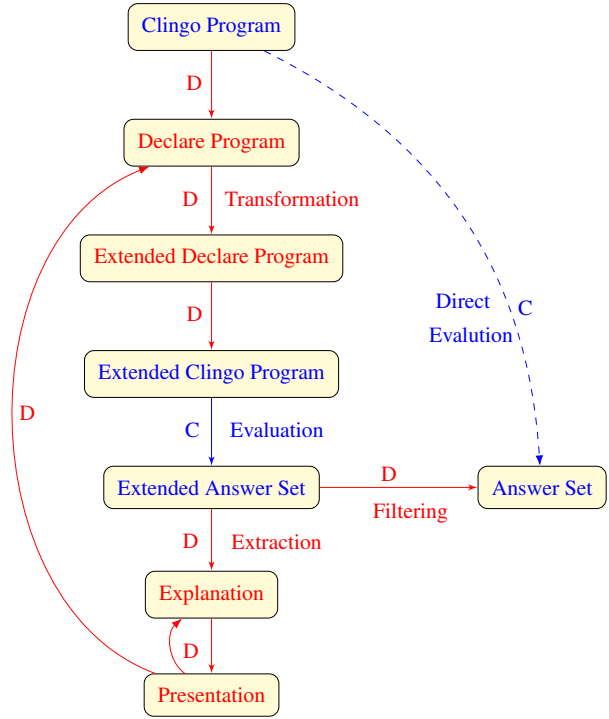
\begin{figure}[htb]
\centering
\vspace*{-0cm}
\begin{center}
\resizebox{.7\textwidth}{!}{\input{workflowASP.tikz}\unskip}
\end{center}
\caption{Workflow of Evaluation and Visualization using Declare and Clingo.}\label{fig:workflow}
\vspace*{-.0cm}
\end{figure}

It is important to note, that we support an incremental explanation generation
process. If the explanations are too complicated, or not sufficient, then we
can refine the explanation, either by modifying the Declare program, or by
extending/reducing the explanation given the answer set. This is indicated in
Figure~\ref{fig:workflow} by the feedback loop from \emph{Presentation} to the
\emph{Declare Program}.

Furthermore, regarding presentation we can in principle provide the explanation
in various form, \cf~\cite{AR:10a} for different explanation and presentation
dimensions. However, one typical option is to use visualization, \ie to
potentially complement the textual justification by a visual representation in
the form of a tree structure showing the dependencies of the facts and fired
rules.

\subsection{Program Transformation using Rule Instrumentation}

Program transformations have also been used in
\cite{seipel1997characterization} for computing the partial
stable models of disjunctive logic programs with a theorem prover
for stable models,
which was specialized to normal logic programs and augmented with
some further features in \cite{janhunen2006unfolding},
and in the well-known magic sets method for deductive databases,
cf.\ \cite{minker2014logic}.

Our program transformation can be applied to Clingo rules with
a head $A$, the positive body atoms $B_1, \ldots, B_n$,
and the test body atoms $C_1, \ldots, C_m$ given as follows:
\begin{quote}
   \( A\ \mbox{:-}\ B_1, \ldots, B_n,\ C_1, \ldots, C_m. \)
\end{quote}
The test body atoms include all atoms which occur under meta-predicates
such as default negation \codett{not} or \codett{count}.
We assume, that this is rule number $j$ of the Clingo program.
The rule can contain variable symbols under certain
conditions, which can be found in the Clingo literature,
e.g.~\cite{DBLP:journals/corr/GebserKKS14}.
Assuming that the variable symbols occuring in the head and the
positive body of the rule are $X_1, \ldots, X_k$, then a further rule
is constructed for recording that the rule has fired:
\begin{quote}
   \( \mathit{rule\_fired}(j, X_1, \ldots, X_k)\ \mbox{:-}\
         A,\ B_1, \ldots, B_n,\ C_1, \ldots, C_m. \)
\end{quote}
Observe, that the head atom $A$ of the original Clingo rule
becomes part of the body of the transformed Clingo rule.
If $\cal P$ was the original Clingo program,
and ${\cal P}_{\mathit{fired}}$ consists of the transformed Clingo
rules, then we later evaluate the extended Clingo program
   \( {\cal P}_{\mathit{ext}} = {\cal P} \cup {\cal P}_{\mathit{fired}} \)
containing the original rules together with the transformed rules.
We assume that the predicate symbol $\mathit{rule\_fired}$ does not
occur in $\cal P$.
Every answer set $I_{\mathit{ext}}$ of ${\cal P}_{\mathit{ext}}$
corresponds to an answer set $I$ of $\cal P$ extended by atoms
for $\mathit{rule\_fired}$ indicating the ground instances
of the rules that have fired.

E.g., assume that the following Clingo rule is rule number 6
with 2 positive body atoms followed by 3 test body atoms:
\begin{quote}
\begin{lstlisting}
cn_lp(Y, Z) :-
   node(Y), node(Z),
   not edge(Y, Z), Y!=Z, n=#count{X:c(X, Y, Z)}.
\end{lstlisting}
\end{quote}
The positive body $B_1, B_2$ is {\tt node(Y), node(Z)},
and the test body atoms $C_1, \ldots, C_3$ are
{\tt not edge(Y, Z), Y!=Z, n=\#count\{X:c(X, Y, Z)\}}.
For given bindings of {\tt Y, Z},
the test atom {\tt n=\#count\{X:c(X, Y, Z)\}} counts the
number of constants~{\tt X}, for which {\tt c(X, Y, Z)}
is true.
Thus, the transformed rule is
\begin{quote}
\begin{lstlisting}
rule_fired(6,Y,Z) :-
   cn_lp(Y, Z), node(Y), node(Z),
   not edge(Y, Z), Y!=Z, n=#count{X:c(X, Y, Z)}.
\end{lstlisting}
\end{quote}
Observe, that the variable symbol $X$ is not part of the
transformed rule, since it does not occur in the head or the
positive body of the original Clingo rule.

Our approach is similar to the offline justifications
of \cite{fandinno2019answering}.
In addition, we can handle non-ground rules with variable symbols,
and we can return the explanations and select a suitable subset
for graphical presentation and visualization.

\subsection{Generating Explanations}

From the atoms $\mathit{rule\_fired}(j, x_1, \ldots, x_k)$ of the
answer sets for the extended Clingo program ${\cal P}_{\mathit{ext}}$,
it can be inferred which rule instances of the original Clingo
program $\cal P$ have fired to support the answer set.
For every derived atom $\mathit{rule\_fired}(j, x_1, \ldots, x_k)$,
where $j$ is a number and $x_1, \ldots, x_k$ are constants,
we know that the $j$--th rule of the original Clingo program $\cal P$
fired with the instantiations $X_i = x_i$ of the variable symbols
occuring in the head or the positive body.
From the substitution
   \( \theta = \{\, X_i \mapsto x_i\, |\, 1 \leq i \leq k\, \} \)
given by the derived atom $\mathit{rule\_fired}(j, x_1, \ldots, x_k)$,
we know that the head atom $A \theta$ is also in the answer set,
and that it is an element of the answer set of $\cal P$.
We can extract
\begin{quote}
   \( (A \theta) - \mathit{is\_supported\_by} -
      (A \theta \mbox{:-}\
       B_1\theta, \ldots, B_n\theta,\ C_1\theta, \ldots, C_m\theta) \)
\end{quote}
as an {\it explanation} of $A \theta$.
If {\tt rule_fired(7,1,3)} was derived in the example above,
then the following instance of rule {\tt 7} of the original
Clingo program has fired to support {\tt cn\_lp(1,3)}:
\begin{quote}
\begin{lstlisting}
cn_lp(1, 3) :- node(1), node(3),
   not edge(1, 3), 1!=3, n=#count{X:c(X, 1, 3)}.
\end{lstlisting}
\end{quote}
The variable symbol $X$ is not bound, since it only occurs in the
test atoms.
We construct the explanation
\begin{quote}
\begin{lstlisting}
cn_lp(1, 3) - is_supported_by -
   ( cn_lp(1, 3) :- node(1), node(3),
        not edge(1, 3), 1!=3,
        n=#count{X:c(X, 1, 3)} ).
\end{lstlisting}
\end{quote}

\subsection{Presenting Explanations}

The number of constructed explanations can be very large,
as turned out in the case studies of Section~\ref{sec:case:study}.
For a suitable visualization, we can query the set of explanations;
e.g., we can select certain explanations
   \( (A \theta) - \mathit{is\_supported\_by} - rule\_instance, \)
such that the predicate symbol of the head $A$ is in a suitable
subset.
Then, the explanations could also be visualized by dependency
graphs.

\section{Case Study: ASP-Based Link Analysis}
\label{sec:case:study}

In the following, we outline our method for link prediction using ASP. The main
strength of ASP is its intuitive way to state a problem, also allowing to scale
the problem up easily, and the availability of computationally powerful ASP
solvers. For this study, the former two points are more relevant since for our
application we utilize a relatively small data set in our case study.
As an ASP solver, we use Clingo \cite{DBLP:journals/corr/GebserKKS14}
embedded in Python.
The following example programs are stratified; \ie there is no
recursion through test atoms (e.g., negated atoms or count atoms).
In more complicated examples, e.g., including more declarative
background knowledge, however, recursion could be used.

\begin{figure}
\centering
{\includegraphics[width=0.4\textwidth]{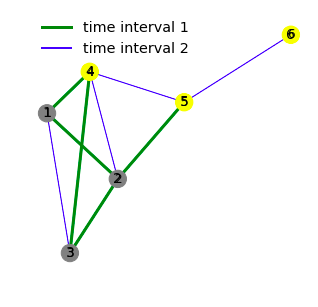}}
\vspace*{-0.5cm}
\caption{Example interaction network represented as an interaction graph: Edges denoting links between actors (as nodes) are split into subsets based on two different time intervals.}
\label{fig:network:example}
\end{figure}

As introduced above, link prediction aims at inferring the (future) link
structure of a network, \eg based on time intervals, such that the network used
for prediction uses the links contained in the first time interval $T_1$ while the
network used for testing uses the links contained in the second (subsequent)
time interval $T_2$, \cf Figure~\ref{fig:network:example}.

As a simple example, the graph $G$ shown in Figure \ref{fig:network:example}
 represents interaction between actors at an event, split into two time frames.
Then, we utilize the information given in time interval 1, \ie the green edges,
in order to infer the link structure in time interval 2, \ie the violet edges.
In particular, we aim at predicting those links given the information on time
interval 1.

A quite simple but usually quite effective approach for link prediction is to
consider the number of common neighbors of two nodes as a ranking for
predicting a link. In the following, we will utilize that simple common
neighbor predictor in an ASP implementation. Besides a simple graph $G_1$, we
can also extend the approach to a bipartite graph $A$ used as background information,
for predicting links, see~Figure~\ref{fig:1} for an example.
Rules can be created in such a way that for a constant $n$ (a threshold), where
$\Gamma_{G}(x)$ stands for the neighborhood of node $x$ in  a graph $G$, $E_{2pred}$
stands for the predicted edges for $T_2$:
\begin{quote}
   \( \forall u, v \in V \mid (u, v) \not\in E_{1},\
      |\Gamma_{A}(u)\cap \Gamma_{A}(v)|=n \implies \\
      \forall x \in \Gamma_{G_1}(u) \backslash \Gamma_{G_1}(v) \mid
      (x, v) \in E_{2pred}\ \land
      \forall y \in \Gamma_{G_1}(v) \backslash \Gamma_{G_1}(u) \mid
      (x, u) \in E_{2pred} \)
\end{quote}
Then, given $G_1$ and $A$, we can predict links for time interval 2,
denoted as $E_{2pred}$.

\subsection{Example: Link Prediction}
\label{sec:didacticexample}

\begin{figure}[htb]
\vspace*{-.0cm}
\begin{center}
\includegraphics[width=8.8cm]{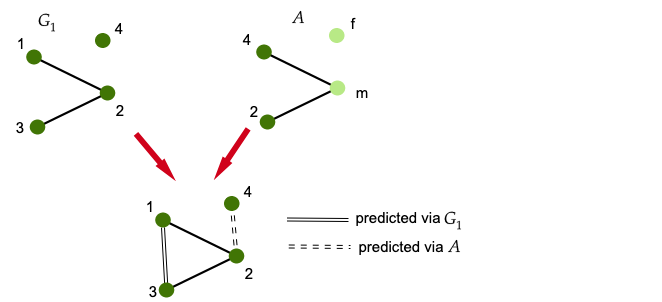}
\end{center}
\caption{Example: Interaction graph $G_1$, and attributive graph $A$, where $f$ and $m$ denote attributes.}
\label{fig:1}
\vspace*{-.0cm}
\end{figure}

Below, we will first  illustrate our approach for link prediction via a small
hypothetical example. The example data is visualized in Figure~\ref{fig:1}.

The Clingo code corresponding to the example in Figure~\ref{fig:1}
describes two symmetric graphs,
the interaction graph and the attributive graph, by giving
the edges and some symmetry rules.
The computed answer set compares the predicted links with
some expected test links.
The first part of the Clingo program defines two constants
and the nodes and edges of the graphs:
\begin{quote} \footnotesize
\begin{lstlisting}
#const n=1.
#const n_attrib=1.

% Interaction Graph: Nodes
node(1..4).
% Edges, first time interval
edge(1, 2).  edge(2, 3). 
% Edges, second time interval (test set)
test(1,3).  test(2,4). 

% Attributive Graph: Nodes and Edges
node_attrib(2).  node_attrib(4).
node_attrib(f).  node_attrib(m).
edge_attrib(2, m).  edge_attrib(4, m). 

% The edge relations are made symmetric,
% such that they correspond to undirected graphs.
edge(Y, X) :- edge(X, Y).
edge_attrib(Y, X) :- edge_attrib(X, Y).
test(Y, X) :- test(X, Y).
\end{lstlisting}
\end{quote}
The link prediction is accomplished by the second part of the
Clingo program:
\begin{quote} \footnotesize
\begin{lstlisting}
% X is a common neighbor of the unconnected Y and Z.
c(X, Y, Z) :- edge(X, Y), edge(X, Z),
   not edge(Y, Z), Y!=Z. 
c_attrib(X, Y, Z) :-
   edge_attrib(X, Y), edge_attrib(X, Z),
   not edge_attrib(Y, Z), Y!=Z.
% A link is predicted, when there is one common neighbor
% in the interaction/attributive graph (c/c_attrib).
cn_lp(Y, Z) :- node(Y), node(Z),
   not edge(Y, Z), Y!=Z,
   n=#count{X:c(X, Y, Z)}.
cn_lp(Y, Z) :- node(Y), node(Z),
   not edge(Y, Z), Y!=Z,
   n_attrib=#count{X:c_attrib(X, Y, Z)}.

% match compares the predicted links with the test links
match(X, Y) :- test(X, Y), cn_lp(X, Y).

#show cn_lp/2.  #show match/2.  
\end{lstlisting}
\end{quote}

In the following, we show a part of the computed answer set,
as requested by the \verb+show+ statements in the clingo code.
The facts for \verb+match+ show that the predicted links
(predicate \verb+cn_lp+) exactly match the expected links
(predicate \verb+test+) of the test set.
\begin{quote}
\begin{lstlisting}
cn_lp(1,3) cn_lp(3,1) cn_lp(2,4) cn_lp(4,2)
match(1,3) match(3,1) match(2,4) match(4,2)
\end{lstlisting}
\end{quote}

The additionally computed facts of the answer set of the extended
clingo program show that
   {\tt cn_lp(2,4)} and {\tt cn_lp(4,2)}
were generated from rule {\tt 6} (line 1 below), whereas
   {\tt cn_lp(1,3)} and {\tt cn_lp(3,1)} were
generated from rule {\tt 7} (line 2 below).
The {\tt match} facts were all generated from rule {\tt 14}
(lines 3 and 4 below).
\begin{quote}
\begin{lstlisting}
rule_fired(6,2,4)   rule_fired(6,4,2)
rule_fired(7,1,3)   rule_fired(7,3,1)
rule_fired(14,1,3)  rule_fired(14,3,1)
rule_fired(14,2,4)  rule_fired(14,4,2)
\end{lstlisting}
\end{quote}

For generating the explanations, we then apply our proposed method.
This results in the following explanations indicating two ground
instances of rule {\tt 7} deriving the facts
{\tt cn_lp(1,3)} and {\tt cn_lp(3,1)}, respectively,
and two ground instances of rule {\tt 14} deriving the facts 
{\tt match(1,3)} and {\tt match(3,1)}, respectively:
\begin{quote}
\begin{lstlisting}
cn_lp(1,3)-is_supported_by-
   ([cn_lp(1,3)]-[node(1),node(3)]-
    [not edge(1,3),1!=3,n= #count{X:c(X,1,3)}])
cn_lp(3,1)-is_supported_by-
   ([cn_lp(3,1)]-[node(3),node(1)]-
    [not edge(3,1),3!=1,n= #count{X:c(X,3,1)}])
match(1,3)-is_supported_by-
   ([match(1,3)]-[test(1,3),cn_lp(1,3)]-[])
match(3,1)-is_supported_by-
   ([match(3,1)]-[test(3,1),cn_lp(3,1)]-[])
\end{lstlisting}
\end{quote}
The analogous explanations of {\tt cn_lp(2,4)} and
{\tt cn_lp(4,2)} are not shown.

\begin{figure}[htb]
\centering
{\includegraphics[width=\textwidth]{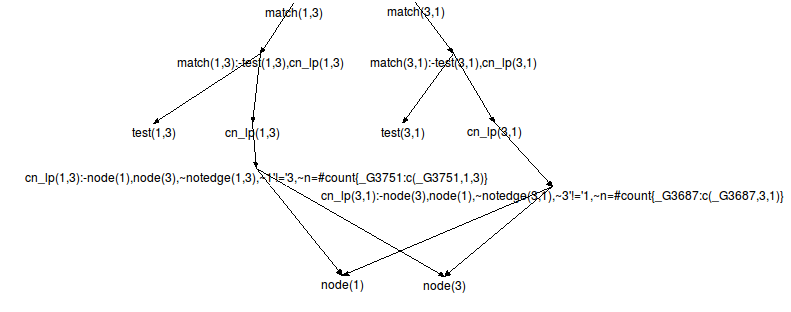}}%
\caption{Example: Presentation of the generated explanation.
   The justification of the answer set is represented
   by a tree structure.}
\label{fig:network:example:explanation}
\end{figure}

For presentation, we show the computed answer set $I$ and we visualize
its explanation, \ie its justification by a tree structure
$\cal T$;
see Figure~\ref{fig:network:example:explanation}.
All body atoms are relevant for the explanations;
the positive body atoms lead to subtrees, whereas the test body atoms
have only been tested in $I$
(which was assured by the atoms for $\mathit{rule\_fired}$).

\subsection{Example: Anomalous Link Discovery}
\label{sec:example:2}

This section describes a more complex example. Here, we are not only interested
in the link prediction, but also in anomalous link discovery. As discussed
above, anomalous link discovery aims at discovering anomalous links that
are in contrast to a given (reference) model~\cite{CA:2019}. In our example, we can
consider a given network, as a reference model for interaction behavior, which
we can then compare to the predicted interaction behavior. Using declarative
programming and particularly answer set programming, it is very simple to
implement that.

\begin{figure}[htb]
  \centering
  \includegraphics[width=.87\columnwidth]{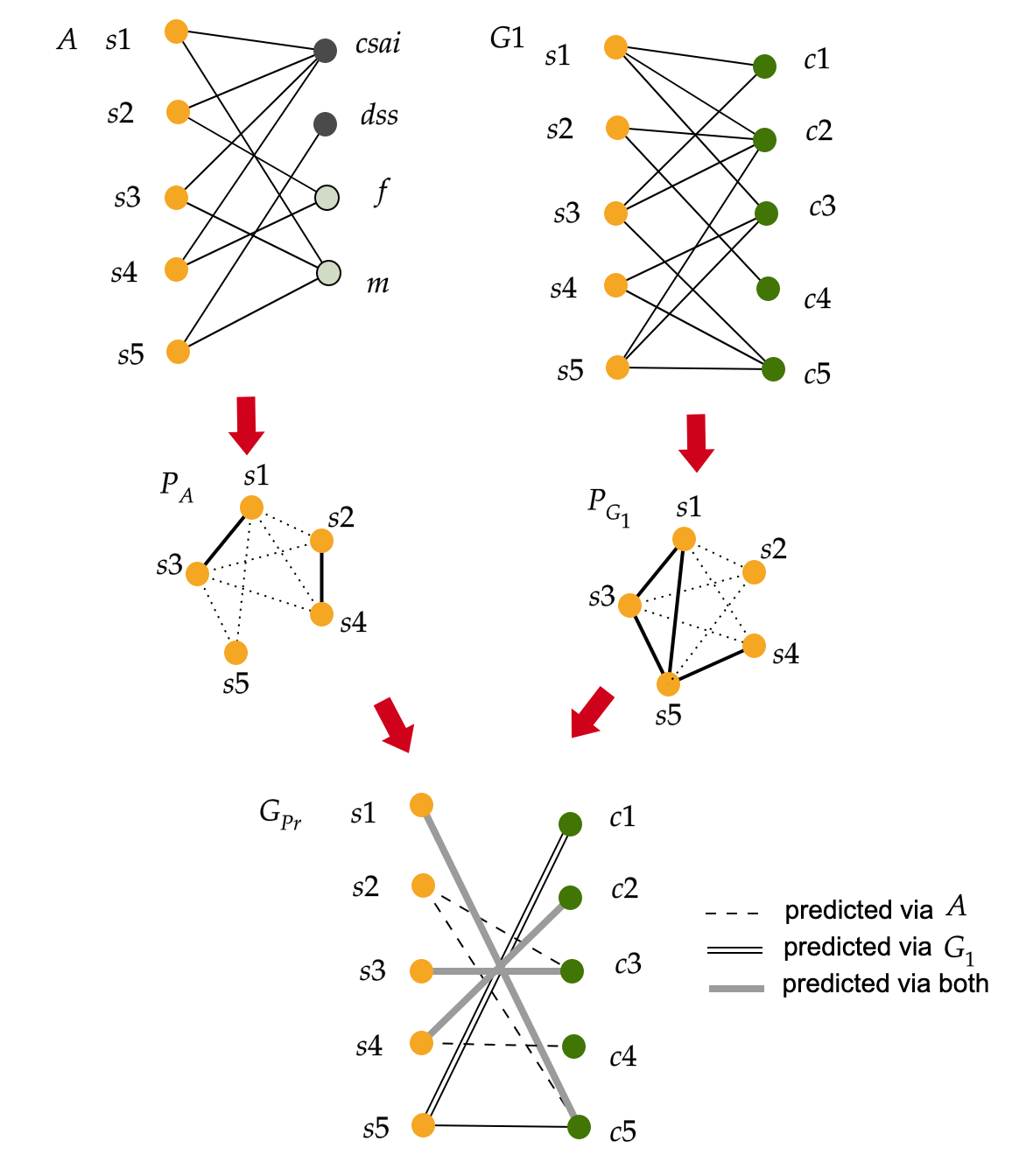}
  \caption{Comparing links predicted via background knowledge ($A$)
     and past links ($G_1$).} \label{fig:comparison}
\end{figure}

First, we demonstrate a case where domain knowledge formalized in a network is
used together with past interactions to predict future links, for a bipartite
graph. 
Consider Figure~\ref{fig:comparison}:
The bipartite graph $A$
represents the choices of the attributive
information provided as background knowledge. The nodes $s_1, s_2, s_3, s_4, s_5$
represents students,  and the nodes $f, m$ represent their gender ($f$: female,
$m$: male). The nodes $dsbg,  csai$  are standing for the master programs the
students are enrolled to, \eg ``Cognitive Science and Artificial Intelligence"
or ``Data Science for Business and Governance".
 
Above, we have described how to connect similar disconnected vertices, where
similarity was determined by the number of common neighbors.
Here, similarity is defined the same way, but we do not connect similar
disconnected vertices;
we predict new links for each based on the neighborhood of the other.

In Figure \ref{fig:comparison},  a pair of bipartite graphs are shown on the top row.
The graph $A$ has a set of vertices partitioned into students $\{s_1,\cdots,
s_5\}$, and a set of  attributes represented by nodes $\{csai, dss, f,m\}$.
Graph $G_1$ has vertices partitioned into students
   $\{s_1,\cdots, s_5\}$,
and companies
   $\{c_1,\cdots, c_5\}$.
This graph captures only the interactions between the students and companies.
The intuition is that similar students have similar interests.
In the second row, we see the projections of the graphs on students ($P_A, P_{G_{1}}$).
A projection graph for a bipartite graph is a weighted graph on the set of
vertices on one of the partitions, where the weight of the edges is determined
by the number of common neighbors.
We highlight the edges where the weight is 2, and assume the students these
edges are joining as similar. Then finally, in the last row we see the graph
showing the predicted links, between the students and the companies. For both
graphs $A,G_1$, for a pair of similar vertices $s_i,s_j$, ($i,j
\in\{1,2,3,4,5\}, i\neq j$) we predict a link between   $s_i,c_k$ if
$(s_j,c_k)$ is an edge in $G_1$, and $(s_i,c_k)$ is not. These two similarity
measures based on two different graphs $A, G_1$ imply different sets of
predicted links say $G_{(Pr,A)}$, $G_{(Pr,G_1)}$.

The ASP code below calculates the predicted links as well as
distinguishes the sets
   \( G_{(Pr,A)} \backslash G_{(Pr,G_1)}, \)
   \( G_{(Pr,G_1)} \backslash G_{(Pr,A)} \)
and
   \( G_{(Pr,G_1)}\cap G_{(Pr,A)} \)
for anomaly analysis purposes.
We omit the first part of the Clingo file. Constants are given as: {\tt n=2, n_attrib=2} (2 common neighbors, see below).
Facts for the networks/graphs are defined as follows:
\begin{itemize}
   \item {\tt node_s(X)}: the 5 yellow nodes of $G_1$,
   \item {\tt edges(X,Y)}: the 13 edges of $G_1$,
   \item {\tt edge_attrib(X,Y)}: the 8 edges of $A$.
\end{itemize}
The graphs are undirected, which is ensured by additional rules
for the edge relations {\tt edge/2} and {\tt edge\_attrib/2}.
X is a common neighbor of Y and Z:
\begin{quote} \footnotesize
\begin{lstlisting}
c(X, Y, Z) :- edge(X, Y), edge(X, Z),
   node_s(X), node_s(Y), node_s(Z), Y!=Z. 
c_attrib(X, Y, Z) :-
   edge_attrib(X, Y), edge_attrib(X, Z),
   node_s_attrib(X), node_s_attrib(Y),
   node_s_attrib(Z), Y!=Z.
\end{lstlisting}
\end{quote}
Two students are similar of type 1/2 ({\tt similar_1/2}, {\tt similar_2/2}),
if they have two common neighbors in
the interaction/attribute graph:
\begin{quote} \footnotesize
\begin{lstlisting}
similar_1(Y, Z) :- node_s(Y), node_s(Z),
   not edge(Y, Z), Y!=Z,
   n=#count{X:c(X, Y, Z)}.
similar_2(Y, Z) :- node_s_attrib(Y), node_s_attrib(Z),
   not edge_attrib(Y, Z), Y!=Z,
   n_attrib=#count{X:c_attrib(X, Y, Z)}.
\end{lstlisting}
\end{quote}
We consider pairs of similar vertices and the nodes
that are adjacent to one but not the other and predict
a link based on that.
\begin{quote} \footnotesize
\begin{lstlisting}
lp_1(X,Y) :- similar_1(Y,Z), edge(X,Z), not edge(X,Y).
lp_2(X,Y) :- similar_2(Y,Z), edge(X,Z), not edge(X,Y).

common(X,Y)  :- lp_1(X,Y), lp_2(X,Y).
diff_12(X,Y) :- lp_1(X,Y), not common(X,Y).
diff_21(X,Y) :- lp_2(X,Y), not common(X,Y).
\end{lstlisting}
\end{quote}
From the Clingo program, the following answer set is computed:

\begin{quote} \footnotesize
\begin{lstlisting}
common(c3,s3) common(c5,s1) common(c2,s4)
diff_12(c1,s5) diff_21(c4,s4)
diff_21(c5,s2) diff_21(c3,s2)
lp_1(c1,s5) lp_1(c3,s3) lp_1(c5,s1) lp_1(c2,s4)
lp_2(c3,s3) lp_2(c4,s4) lp_2(c2,s4)
lp_2(c5,s1) lp_2(c5,s2) lp_2(c3,s2)
\end{lstlisting}
\end{quote}
When we perform our proposed method, selecting {\em diff\_12}, {\em diff\_21}, {\em lp\_1} and
{\em lp\_2} as predicates, we obtain an extended answer set of size 144, which
already indicates the (large) size of the explanation. When generating the
explanation on this extended answer set, we obtain a set of size 136
accordingly. This already shows the complexity, since a naive visualization
approach for presentation would result in a presentation that is not
understandable. Therefore, according selection on specific instances and
predicates needs to be performed by the user, in an incremental fashion.

\section{Conclusions}
\label{sec:conclusions}

In this paper, we have proposed a novel method for generating explanations in
an incremental process using declarative program transformations. Since the
method itself is purely based on syntactic transformations of declarative
programs, the method enables in principle a general approach. However, in the
context of this paper we exemplified the approach by focusing on answer set
programming -- as a prominent tool for declarative programming. Our
implementation itself is embedded into the Declare software toolkit. For the
program evaluation we directly connect to Clingo.
We have exemplified the application and efficacy of the proposed approach in the
context of link analysis, \ie link prediction and anomalous link discovery for
social networks. Our exemplary results indicate that the method performs well
for obtaining explanations and according presentations which can also be
incrementally refined.

For future work, we aim to extend the approach towards knowledge-based
automatic refinement methods, also taking into account more complex (and
richer) data representations, \eg in the field of complex interaction
networks~\cite{IAGKLS:19}.

\section*{Acknowledgments}

This work has been partially supported by the German Research Foundation (DFG)
under grant AT 88/4-1.


\end{document}

%% file: workflowASP.tikz
\begin{tikzpicture}[node distance = 2cm, auto]

\node [wa,text=blue] (CL_Pr) {Clingo Program};
\node [ text=red,below left=.25cm and -1.2cm of CL_Pr] (Declare) {D};
\node [wa, text=red, below=.99cm of CL_Pr] (DL_Pr) {Declare Program};
\node [wa, text=red, below=.99cm of DL_Pr] (DL_Pr2) {Extended Declare Program};
\node [ text=red,below left=.25cm and -1.2cm of DL_Pr] (Declare2) {D};
\node [ text=red ,below right=.25cm and -1.2cm of DL_Pr] (Tr) {{Transformation}};
\node [ text=red,below left=.25cm and -1.9cm of DL_Pr2] (Declare3) {D};
\node [wa, text=blue, below=.99cm of DL_Pr2] (CL_Pr2) {Extended Clingo Program};

\node [ text=blue,below left=.25cm and -1.8cm of CL_Pr2] (Clingo) {C};
\node [ text=blue, below right=.25cm and -1.7cm of CL_Pr2] (Clingo_Eval) {Evaluation};

\node [ text=blue, above right=0.4cm and 2.5cm of CL_Pr2] (Clingo_Direct) {C};
\node [ text=blue, above right=.5cm and 1.3cm of CL_Pr2] (Clingo_Direct_Text) {Direct};
\node [ text=blue, above right=0.0cm and 1.2cm of CL_Pr2] (Clingo_Evaluation_Text) {Evalution};

\node [ text=red, below left= 0.0cm and .6cm of CL_Pr2] (Declare_Iteration) {D};

\node [wa, text=blue, below=.99cm of CL_Pr2] (AS) {Extended Answer Set};
\node [wa, text=blue, right=2.3cm  of AS] (Rs) {Answer Set};
\node [ text=red,below right=.9cm and 0.5cm of CL_Pr2] (Declare4) {D};
\node [ text=red, below left=.1cm and -1.2cm of Declare4] (ExtR) {Filtering};


\node [wa, text=red, below =.9cm  of AS] (Exp) {Explanation};
\node [ text=red, below left=.25cm and -1.5cm of  AS] (Declare5) {D};
\node [ text=red, below right=.25cm and -1.4cm of AS] (ExtR) {Extraction};
\node [ text=red,below left=.25cm and -.9cm of Exp] (Declare6) {D};

\node [wa, text=red, below=.8cm of Exp] (Vis) {Presentation};
\draw [-{Latex[length=1.5mm]},blue,dashed] (CL_Pr) to [bend left=30]  node [above, sloped]  (TestNode1) {} (Rs);

\draw [-{Latex[length=1.5mm]},red] (Vis) to [bend left=51]  node [above, sloped]  (TestNode2) {} (Exp);
\draw [-{Latex[length=1.5mm]},red] (Vis) to [bend left=69]  node [above, sloped]  (TestNode2) {} (DL_Pr);

\path [line,red] (CL_Pr) -- (DL_Pr);
\path [line,red] (DL_Pr) -- (DL_Pr2);
\path [line,red] (DL_Pr2) -- (CL_Pr2);
\path [line,blue] (CL_Pr2) -- (AS);
\path [line,red] (AS) -- (Exp);
\path [line,red] (AS) -- (Rs);
\path [line,red] (Exp) -- (Vis);


\end{tikzpicture}